\documentclass[final,3p,times,twocolumn]{elsarticle}

\usepackage{amssymb}
\usepackage{amsmath}

\usepackage{algorithm}
\usepackage{newfloat}
\usepackage{listings}
\usepackage{booktabs}
\usepackage{multirow}
\usepackage{amssymb}
\usepackage{amsmath}
\usepackage{float}
\usepackage{algorithmic}
\usepackage{xcolor}
\usepackage{soul}
\usepackage{caption}

\journal{Knowledge-Based Systems}

\begin{document}

\definecolor{mypink0}{rgb}{0.95,0.78,0.96}
\definecolor{mygreen0}{rgb}{0.83,0.93,0.77}
\definecolor{myyellow0}{rgb}{0.98,0.93,0.72}
\definecolor{myblue0}{rgb}{0.67,0.85,1.00}
\definecolor{mypurple0}{rgb}{0.87,0.80,0.92}









\soulregister{\ref}{1}
\soulregister{\cite}{1}
\soulregister{\textbf}{1}
\soulregister{\emph}{1}
\soulregister{\leq}{0}  

\soulregister{\hl}{1} 
\newcommand{\Rone}[1]{\begingroup\sethlcolor{mypurple0}\hl{#1}\endgroup}
\newcommand{\Rtwo}[1]{\begingroup\sethlcolor{mygreen0}\hl{#1}\endgroup}




\begin{frontmatter}



\title{MindShot: A Few-Shot Brain Decoding Framework via Transferring Cross-Subject Prior and Distilling Frequency Domain Knowledge} 

\author[label1]{Shuai Jiang} 
\ead{js@bupt.edu.cn}
\author[label1]{Zhu Meng} 
\ead{bamboo@bupt.edu.cn}
\author[label1]{Haiwen Li} 
\ead{lihaiwen@bupt.edu.cn}
\author[label1]{Delong Liu} 
\ead{liudelong@bupt.edu.cn}
\author[label1,label2]{Fei Su} 
\ead{sufei@bupt.edu.cn}
\author[label1,label2]{Zhicheng Zhao\corref{corresponding}} 
\ead{zhaozc@bupt.edu.cn}
\cortext[corresponding]{Corresponding author.}


\affiliation[label1]{organization={Beijing University of Posts and Telecommunications},
            city={Beijing},
            country={China}
            }
\affiliation[label2]{organization={Beijing Key Laboratory of Network System and Network
Culture},
            city={Beijing},
            country={China}
            }

\begin{abstract}
Aiming to reconstruct visual stimuli from brain signals, brain decoding has recently made significant progress using functional magnetic resonance imaging (fMRI). However, it still has challenging issues such as substantial individual differences and high data collection costs. To simplify these problems, most methods adopt the per-subject-per-model paradigm, but this greatly limits their applications. In this paper, we design a few-shot brain decoding setting specifically for potential clinical scenarios and propose a novel two-stage decoding framework named MindShot, comprising a Multi-Subject Pretraining (MSP) stage and Fourier-based cross-subject Knowledge Distillation (FKD) stage. Firstly, a MSP framework based on multi-modal contrastive learning is constructed to mine the cross-subject prior. Secondly, the FKD is presented to decrease inter-individual differences while improving the decoding adaptability to new individuals. Our approach achieves high semantic fidelity in visual reconstruction on the largest dataset and has the potential to reduce scanning time by up to 99\%. Remarkably, MindShot achieves a CLIP accuracy of 83.6\% using only 1.8\% of the fMRI-image pairs, surpassing the 77.4\% accuracy of the method trained on the entire NSD dataset. This makes it feasible to train large-scale brain decoding frameworks that require less data, facilitating practical applications. The code is available at https://github.com/JSinBUPT/MindShot.

\end{abstract}




\end{frontmatter}



\section{Introduction}\label{Introduction}

The brain, as the core of human cognition and perception, plays a pivotal role in encoding and processing perceptual stimuli, as well as influencing decisions. A deeper understanding of the brain's mechanisms requires exploring the reverse process---brain decoding---which involves interpreting responses from external visual stimuli to the visual cortex \cite{cox2003functional,kamitani2005decoding}, and reconstructing images based on functional magnetic resonance imaging (fMRI) has become a feasible solution \cite{naselaris2009bayesian}. 

\begin{figure}[t]
    \centering
    \includegraphics[width=1\linewidth]{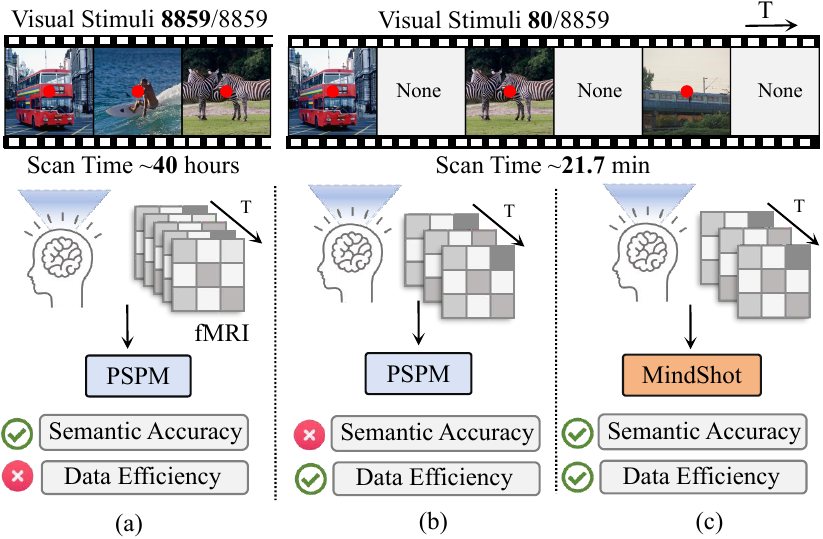}
    \caption{Comparison with existing brain decoding methods. (a) Current fMRI-to-image methods rely on training PSPM paradigms with approximately 40 hours of scan data. (b) These methods suffer from significant performance degradation when only limited data is available. (c) Our MindShot demonstrates improved performance by effectively leveraging cross-subject prior knowledge in few-shot scenarios. }
    \label{fig: introduction}
\end{figure}

Encouraging progress has been made in brain decoding. From generative adversarial networks (GANs) \cite{goodfellow2020generative,ozcelik2022reconstruction,seeliger2018generative} to diffusion models (DMs) \cite{ takagi2023high,scotti2024reconstructing, lu2023minddiffuser,ozcelik2303brain}, these approaches have enabled the reconstruction of more realistic and semantically faithful images. However, brain decoding still faces many challenges.

Specifically, there are substantial individual differences among subjects, that is, individuals will exhibit unique responses to the same visual stimulus \cite{horikawa2017generic,chen2023seeing}. Therefore, most of the existing methods adopt a per-subject-per-model (PSPM) paradigm \cite{horikawa2017generic,chen2023seeing,seeliger2018generative,takagi2023high,scotti2024reconstructing, lu2023minddiffuser,ozcelik2303brain}, in which the decoding model needs to be trained separately for each subject. When applied to new subjects, brain scans have to be repeated. However, acquiring brain data using fMRI is costly and technically demanding, severely limiting the applicability of the brain decoding model to new individuals. These limitations motivate us to propose a new task: \textbf{few-shot brain decoding}, which aims to achieve visual reconstruction by training on only a small number of fMRI-image pairs from new subjects.

However, achieving few-shot brain decoding is also difficult. As shown in Fig.\ref{fig: introduction} (a) and (b), when sufficient data (approximately 40 h of scan time) are available for new subjects, the PSPM can achieve relatively high semantic accuracy. In contrast, under a few-shot scenario (approximately 21.7 min of scan time), its performance degrades significantly. Recent studies have explored cross-subject frameworks with the potential to reduce the scan time required for new subjects \cite{wang2024mindbridge, scotti2024mindeye2, Liu_Ma_Zhu_Jing_Zheng_2025, Xu_2025_ICCV}. In this paper, we investigate how brain decoders can be transferred across individuals, so as to (1) understand the proportion of the meaningful part signal variability between participants and (2) derive cost-optimal acquisition strategies for building models that may be useful in clinical settings.

Accordingly, we propose MindShot, a novel few-shot brain decoding framework that leverages cross-subject prior knowledge to enhance performance in few-shot scenarios. First, prior knowledge across multiple subjects is captured via multi-modal contrastive learning. Next, the brain decoding model is fine-tuned using only a small number of fMRI-image pairs from new subjects. Notably, the lack of effective biological guidance make interpreting the complex neural activity of new subjects from fMRI \cite{chen2023seeing} a significant challenge. To relieve this, we propose the Fourier-based cross-subject Knowledge Distillation (FKD) module, which emphasizes the relevant and meaningful components of the signal in the frequency domain, thereby reducing individual differences and facilitating the transfer of prior knowledge across subjects.

Extensive experiments on the Natural Scenes Dataset (NSD) \cite{allen2022massive} demonstrate that MindShot can reconstruct images with high semantic fidelity while training only a few parameters, and it significantly outperforms the state-of-the-art (SOTA) methods based on the per-subject-per-model paradigm. Additionally, our approach has the potential to reduce scan time by up to 99\%. That is, by using only 1.8\% of the fMRI-image pairs, it achieves performance comparable to methods trained on complete datasets. This result further highlights the advantages of our few-shot brain decoding model.

Our contributions can be summarized as follows:

\begin{itemize}
\item A new few-shot brain decoding task is designed and a few-shot brain decoding framework---MindShot---is accordingly proposed to alleviate the scarcity of new-subject fMRI-image pairs by incorporating cross-subject prior.

\item A Multi-Subject Pretraining (MSP) framework based on multi-modal contrastive learning is constructed to mine the cross-subject prior.
\item  The Fourier-based cross-subject Knowledge Distillation (FKD) module is presented to reduce inter-individual differences, improve the decoding adaptability to new subjects, and ultimately enhance the semantic fidelity of reconstructed images.

\end{itemize}

\section{Related work}

\subsection{Brain decoding}

The development of brain decoding is closely related to the evolution of decoding methods for brain activity. In early years, Shen \textit{et al.} \cite{shen2019deep} design a feature decoder to map fMRI patterns to multi-layer pretrained DNN features of real images. With the emergence of GANs \cite{goodfellow2020generative}, researchers shift the study of fMRI modalities from mapping to DNNs to the latent space of GANs, which facilitates the reconstructing of human faces \cite{du2020structured} and natural images \cite{seeliger2018generative}. 

Recently, brain decoding frameworks have advanced considerably with the development of multimodal technologies, particularly contrastive learning \cite{radford2021learning} and diffusion models (DMs) \cite{rombach2022high}. Different methods map fMRI data to a variety of supervised information, including high-level CLIP semantic embedding \cite{takagi2023high,scotti2024reconstructing, lu2023minddiffuser,ozcelik2303brain}, and low-level depth \cite{takagi2023improving} and color \cite{xia2024dream} features. Then, the encoded brain activity is fed to DMs to synthesize images with semantic fidelity details. Most recently, Wang \textit{et al.} \cite{wang2024mindbridge} has achieved cross-subject brain decoding with a single model using a cyclic fMRI reconstruction mechanism. Ferrante \textit{et al.} \cite{ferrante2024through} utilize ridge regression for multi-subject functional alignment, achieving results by using only approximately 10\% of the total data. Scotti \textit{et al.} \cite{scotti2024mindeye2} introduce shared-subject models, enabling fMRI-to-image decoding with just one hour of data. 

In these representative methods, different network architectures are carefully designed, and the potential of cross-subject frameworks is also explored, thereby promoting brain decoding research and achieving positive progress under the PSPM paradigm. However, existing models still have some limitations, such as insufficient adaptability to new subjects and a heavy reliance on extensive scanning time. In contrast, we propose a cross-subject framework that reduces individual differences through frequency-domain distillation and applies cross-subject knowledge transfer, thereby minimizing the model's reliance on extensive data.


\subsection{Diffusion probabilistic models}

DMs show great potential in multiple applications \cite{blattmann2023stable, yang2023diffusion}. For instance, by leveraging prior knowledge from large-scale image-text pairs \cite{schuhmann2022laion}, DMs demonstrate a strong ability to synthesize high-quality images characterized by high semantic fidelity and resolution. To reduce computational cost, Rombach \textit{et al.} \cite{rombach2022high} employ a variational auto-encoder (VAE) to transfer the denoising process to the latent space. ControlNet \cite{zhang2023adding} aims to enhance the control of the image generation process by fusing other information such as sketch, depth and spatial palette. Versatile Diffusion \cite{xu2023versatile} supports the use of multimodal inputs and outputs. In this paper, we incorporate generative models into our few-shot framework for fMRI-to-image reconstruction.

\subsection{Robust learning from low-quality data} 

In real-world applications, the collected brain decoding datasets often have the “small” nature, that is, they are limited in quantity and accompanied by relatively low quality, such as anomalies, missing values, and noisy labels  \cite{li2024small}. To address this issue, Wang \textit{et al.} \cite{wang2024novel} propose the ETALF method, which combines the self-attention mechanism with an asymmetric loss that adaptively penalizes incorrect categories. This framework improves the model’s robustness when tackling the challenge of Few-shot Fault Diagnosis with Noisy Labels (FFDNL). Zhang \textit{et al.} \cite{zhang2024personalized} introduce a federated learning framework named FedFGCR, which incorporates an adaptive interpolation strategy and a novel regularization term. This design enables each local model to leverage global semantic information while enhancing adaptability to client-specific data distributions. Furthermore, Fang \textit{et al.} \cite{fang2024gfe} propose GFE-Mamba, a multimodal classifier based on a Generative Feature Extractor. The framework can generate PET-like feature representations from MRI data and achieves effective multimodal fusion even without PET scans, thereby improving classification performance. Zhao \textit{et al.} \cite{ZhuShe_Bridging_MICCAI2025} propose a Transformer-based knowledge distillation framework MST-KDNet for brain tumor segmentation under scenarios with missing MRI modalities.

Inspired by these advancements, this paper proposes a few-shot framework that transfers cross-subject priors and distills frequency-domain knowledge to improve model robustness.

\section{Methodology}

\begin{figure*}[h]
    \centering
    \includegraphics[width=1\linewidth]{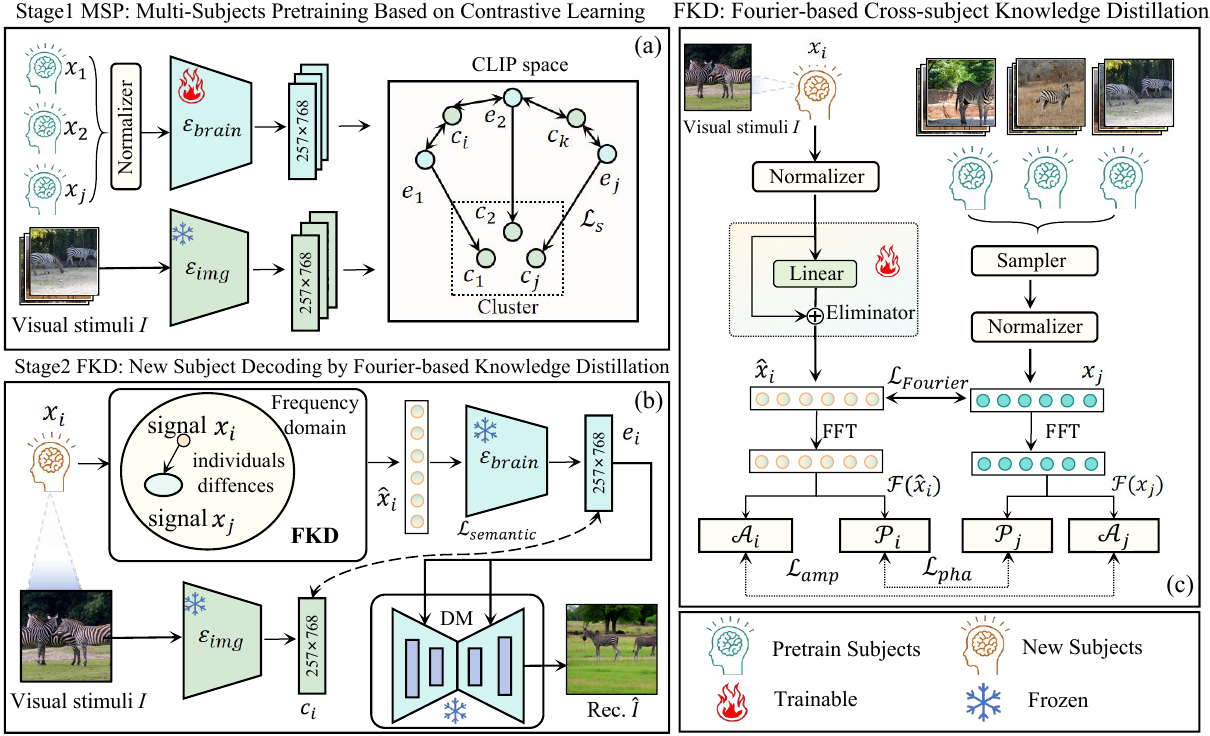}
    \caption{The overall architecture of MindShot. MindShot is a two-stage framework designed for few-shot fMRI-to-image brain decoding. (a) In the first stage, multi-subject prior knowledge pretraining is acquired through a multi-modal contrastive learning using the $\mathcal{L}_{semantic}$, which leads to the pretrained $\mathcal{E}_{brain}$. (b) In the second stage, under the few-shot setting, we propose a Fourier-based cross-subject Knowledge Distillation (FKD) module to eliminate individual differences in the frequency domain, thereby transferring the prior knowledge from $\mathcal{E}_{brain}$ and improving image reconstruction based on diffusion model. (c) The FKD module eliminates individual differences by aligning the dimensions of new and pretrained subjects using a Normalizer first, and then applying an Eliminator to adjust the signal distribution of the new subject, resulting in the signal $\hat{x}_i$. where the $\mathcal{L}_{Fourier}$ is the supervision in the frequency domain. This is achieved by transforming the signals to the frequency domain using FFT, and then minimizing the distances between $\mathcal{A}_i$, $\mathcal{A}_j$, $\mathcal{P}_i$, and $\mathcal{P}_j$.}
    \label{fig:framework}
\end{figure*}

This section presents the proposed few-shot brain decoding framework, which consists of three main parts: the MSP framework, FKD module, and training objectives, as shown in Fig.\ref{fig:framework}. Specifically, in the first stage, MindShot uses multi-modal contrastive learning to acquire cross-subject prior knowledge. In the second stage, few-shot brain decoding for new subjects is achieved by leveraging the proposed FKD module, the pretrained brain encoder, and the diffusion model. Finally, the FKD module mitigates individual differences in the frequency domain by incorporating the Eliminator and the proposed $\mathcal{L}_{Fourier}$ loss.

\subsection{Few-shot brain decoding task}

To clarify the few-shot setting and the overall MindShot framework, we first introduce the Natural Scenes Dataset (NSD) \cite{allen2022massive}, the largest available dataset for brain activity visual reconstruction. NSD contains high-resolution 7T fMRI data from 8 healthy subjects, with a total of 73,000 natural images from the MS-COCO dataset \cite{lin2014microsoft} that are presented to them and categorized into 80 semantic classes. Following prior work \cite{takagi2023high, ozcelik2023natural, scotti2024reconstructing, wang2024mindbridge}, we focus on four subjects (1, 2, 5, and 7), using preprocessed 1D fMRI voxels from the `NSDGeneral' brain regions, with each subject contributing approximately 13,000 to 16,000 voxels. In the proposed few-shot setting, only a small number of samples per semantic class are randomly selected for adaptation. In the 1-shot case, this results in 80 fMRI-image pairs (approximately 21.7 min of scanning). 




\subsection{Multi-subject pretraining (MSP)} \label{Section: Cross-subject prior knowledge pretrain}

In the pretraining stage, it is essential to train a brain encoder $\mathcal{E}_{brain}$ to acquire prior knowledge from multiple subjects. Previous works \cite{takagi2023high, ozcelik2303brain, scotti2024reconstructing} have developed subject-specific brain encoders trained on fMRI-image pairs. However, due to individual differences and differences in regions of interests (ROIs), such models are difficult to apply to cross-subject training. Chen \textit{et al.} \cite{chen2023seeing} propose a masked brain modeling approach that pretrains on multi-subject data and captures the structural patterns of fMRI signals. However, this method incorporates fMRI signals unrelated to visual stimuli, which leads to limited semantic representation.

Hence, we propose a multi-modal contrastive learning approach that enables cross-subject prior knowledge with high semantic alignment. As shown in Fig.\ref{fig:framework} (a), we first use a Normalizer to align the fMRI signals $x_1, x_2, \dots, x_j$ from multiple subjects in response to visual stimuli. The brain encoder $\mathcal{E}_{brain}$ then extracts brain embeddings $e_1, e_2, \dots, e_j$, where $e_j \in \mathbb{R}^{257 \times 768}$. Visual stimuli are encoded by the CLIP image encoder $\mathcal{E}_{img}$ \cite{radford2021learning}, producing image embeddings $c_1, c_2, \dots, c_j$, where $c_j \in \mathbb{R}^{257 \times 768}$. Through contrastive learning, we map brain signals into the CLIP image embedding space, so as to align different individuals who experience semantically similar stimuli. Only the brain encoder $\mathcal{E}_{brain}$ is trained, while $\mathcal{E}_{img}$ remains frozen. In our implementation, the brain encoder $\mathcal{E}_{brain}$ utilizes a multilayer perceptron (MLP) architecture with three adjustable components: a backbone, a projector, and a reconstructor. The projector maximizes semantic similarity between fMRI and image embeddings, while the reconstructor minimizes the spatial distance between the two modalities. 

This architecture enables the learning of shared semantic representations across subjects, thus can enhance the transferability to new individuals. Furthermore, the framework supports pretraining on larger datasets with more diverse subjects, facilitating the acquisition of richer prior knowledge.

\subsection{Fourier-based cross-subject knowledge distillation module} \label{Section: Fourier-based cross-subject module}

In the few-shot brain decoding task, the limited data obtained from new individuals requires identifying the meaningful proportion of signal variability across participants to facilitate an efficient transfer of prior knowledge. Therefore, before inputting fMRI signals into the brain decoder $\mathcal{E}_{brain}$, it is essential to ensure that the statistical distribution of fMRI voxels is aligned with the pretrained multi-subject distribution and to design an effective supervision method.

As illustrated in Fig.\ref{fig:framework} (c), the proposed FKD module first employs a Normalizer to align the new subject's signal $x_i$ with the dimensions of pretrained subjects. The signal $\hat{x}_i$ is produced by an Eliminator which is designed to mitigate inter-subject differences while preserving the original information of the new subject. In our design, the Eliminator adopts a residual-linear structure, where the linear layer predicts individual differences, and the predicted residuals are added back to the original signal. However, due to substantial inter-individual variability, this estimation process is inherently unstable.

The hemodynamic response function (HRF) \cite{lindquist2009modeling} is commonly used to model the relationship between neural activity and the blood oxygenation level dependent (BOLD) signal. In a linear time-invariant (LTI) system, the signal $y(t)$ at time $t$ is considered as the convolution of the stimulus function $s(t)$ and the hemodynamic response $h(t)$, represented as $y(t) = s(t)* h(t)$.
The function $h(t)$ is typically modeled as a linear combination of basis functions, which can be expressed in matrix form as $\mathbf{Y} = \mathbf{X} \mathbf{\beta} +  \mathbf{e}$, where $\mathbf{Y}$ denotes the observed fMRI voxels, $\mathbf{\beta}$ is a vector of regression coefficients, and $\mathbf{e}$ is a vector of unexplained errors. Importantly, the vector $\mathbf{e}$ varies significantly among individuals and across test sessions, since factors such as age and cognitive state can influence firing rate, onset latency, and neuronal activity duration \cite{logothetis2008we} even under the same visual stimuli. This highlights the need for a well-generalized supervision method to guide the Eliminator.

Inspired by previous findings of the Fourier transform, the phase component of the Fourier spectrum preserves high-level semantic information of the signal, whereas the amplitude component contains low-level statistical information, generally regarded as style information \cite{oppenheim1979phase, oppenheim1981importance, piotrowski1982demonstration, hansen2007structural, xu2021fourier, goelman2017frequency,juvells1991role}. In the research related to fMRI, \cite{arja2010changes, goelman2017frequency, yang2018frequency, logothetis2001neurophysiological, goense2008neurophysiology} indicate that the phase captures the sequential neural processing across brain regions and serves as a marker of cognitive information flow, whereas the amplitude is shaped by local neuro-vascular coupling and results in individual and region-specific physiological variations. Correspondingly, we model individual differences in the frequency domain.


Specifically, for a subject's brain signal $x$, its Discrete Fourier Transform (DFT) is formulated as:

\begin{equation}
        \mathcal{F}_{k}(x) = \sum_{n=0}^{N-1} x(n) \mathrm{e}^{-j2\pi \frac{nk}{N}}, k=0,1,...,N-1,    
\label{Eqa: F}
\end{equation}
where $N$ represents the length of the Fourier spectrum signal. This can be efficiently computed using the FFT algorithm \cite{nussbaumer1982fast}. The amplitude and phase components are then represented as:

\begin{equation}
        \mathcal{A}(x) = [R^2(x)+I^2(x)]^{\frac{1}{2}}, \mathcal{P}(x) = arctan[\frac{I(x)}{R(x)}],
\label{Eqa AP}
\end{equation}
where $R(x)$ and $I(x)$ represent the real and imaginary parts of $\mathcal{F}(x)$, respectively.

Next, the Sampler and Normalizer are used to extract semantically consistent visual stimuli and align the dimensions from pretrained subjects. $\mathcal{A}_i$, $\mathcal{A}_j$, $\mathcal{P}_i$, and $\mathcal{P}_j$ are then calculated separately for the signals $\hat{x}_i$ and $x_j$ according to Eq.\ref{Eqa: F} and Eq.\ref{Eqa AP}. The corresponding losses for $\mathcal{L}_{amp}$ and $\mathcal{L}_{pha}$ are computed as follows:

\begin{equation}
        \mathcal{L}_{amp} = \frac{1}{N} \sum_{n=0}^{N-1} (\mathcal{A}_i-\mathcal{A}_j)^2,  
        \mathcal{L}_{pha} = \frac{1}{N} \sum_{n=0}^{N-1} (\mathcal{P}_i-\mathcal{P}_j)^2.
\label{Eqa lamp lpha}
\end{equation}

Finally, the proposed $\mathcal{L}_{Fourier}$ is the weighted sum of $\mathcal{L}_{amp}$ and $\mathcal{L}_{pha}$, and expressed as follows:

\begin{equation} 
\mathcal{L}_{Fourier} = \mathcal{L}_{pha} + \alpha\mathcal{L}_{amp},
\label{Eqa Fourier}
\end{equation}
where the hyperparameter $\alpha$ balances the contributions of the phase and amplitude components. By training the Eliminator with $\mathcal{L}_{Fourier}$, both the high-level semantics and low-level statistical information of the two signals are captured to represent the differences between the individuals.

\begin{figure*}[h]
    \centering
    \includegraphics[width=1\linewidth]{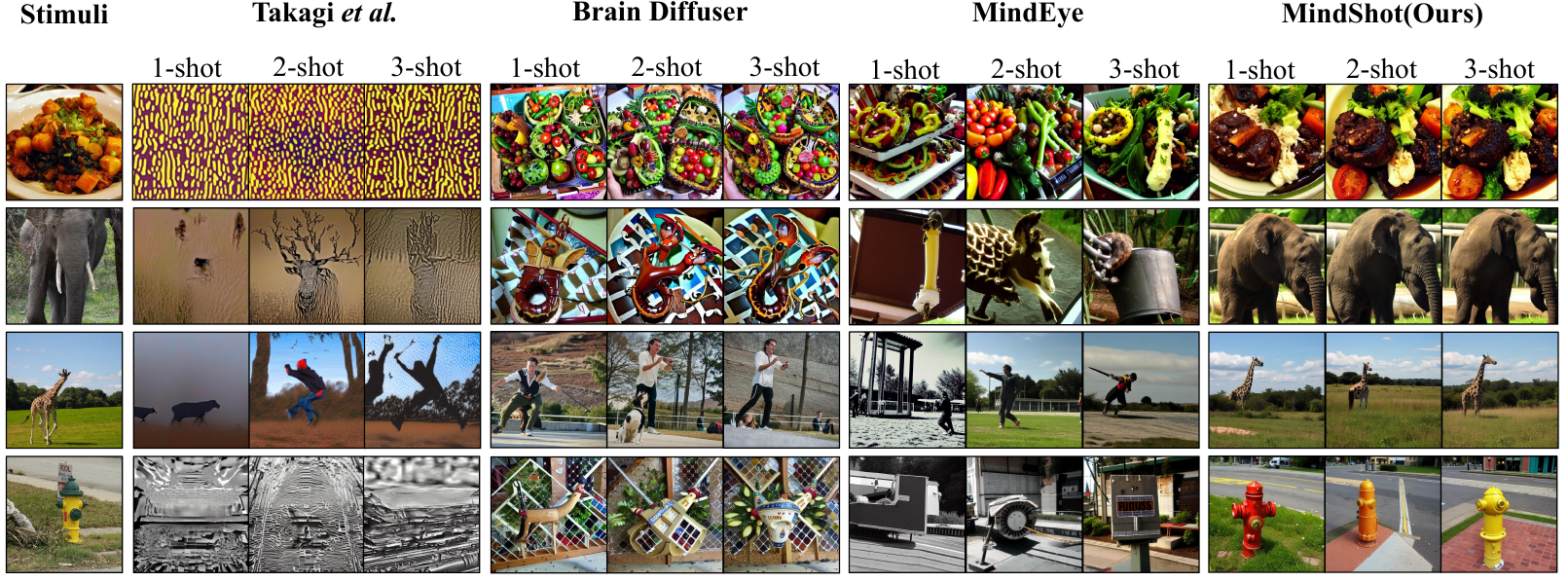}
    \caption{Quantitative comparison with state-of-the-art methods under few-shot settings. Our method reconstructs images with both semantically meaningful content and fine-grained details.}
    \label{fig:visual}
\end{figure*}

\subsection{New subject few-shot brain decoding} \label{Section: New subject few-shot brain decoding}

Considering the limited availability of fMRI-image pairs, we utilize a large-scale pretrained image generation model to assist in image reconstruction. To further ensure semantic fidelity and preserve fine-grained details, we employ image-to-image diffusion models that leverage CLIP image embeddings to guide the image reconstruction. The implementation is as follows:

\begin{equation} \hat{{\epsilon}}_{\theta}(z_t, c) = {\epsilon}_{\theta}(z_t) + s(\theta(z_t, c)-\theta(z_t)), \end{equation}
where $c$ is the conditioning input, $s$ represents the guidance scale, and $z_t$ denotes the latent variable. $\hat{\epsilon}_{\theta}$ is the Unet-based denoising function estimator. The denoising process occurs over $T$ timesteps, yielding the denoised latent variable $z$, which is then decoded via the VAE \cite{kingma2013auto} to generate the image:

\begin{equation} \hat{I} = \mathcal{D}(z). \end{equation}

In the few-shot brain decoding stage of the new subject, as shown in Fig. \ref{fig:framework} (b), we first use the FKD module equipped with an Eliminator and $\mathcal{L}_{Fourier}$, as described in Section \ref{Section: Fourier-based cross-subject module}, to mitigate inter-individual differences in the frequency domain. Then, the pretrained $\mathcal{E}_{brain}$ from Section \ref{Section: Cross-subject prior knowledge pretrain} is leveraged to map the new individual's brain modality signal $\hat{x}_j$ into a CLIP image embedding, which serves as a conditioning input to guide image reconstruction.

\subsection{Learning objectives} \label{Section: Learning objects}

The learning objectives of MindShot involve multi-modal pretraining across subjects, followed by knowledge transfer to new subjects using the FKD module.

In the first stage, the $\mathcal{E}_{brain}$ is trained based on the semantic loss $\mathcal{L}_{semantic}$, which is defined as:
\begin{equation}
        \mathcal{L}_{semantic} = \mathcal{L}_{softCLIP} + \lambda\mathcal{L}_{rec.},
\end{equation}
where the hyperparameter $\lambda$ is used to balance the two losses. The multimodal contrastive learning objective employs the SoftCLIP loss \cite{scotti2024reconstructing} and is defined as follows:
\begin{equation}
\begin{aligned}
    &\qquad \qquad \mathcal{L}_{SoftCLIP}(e,c) =  -\sum_{i=0}^{N-1} \sum_{j=0}^{N-1} \\
    & \frac{\exp(c_i \cdot c_j / \tau)}{\sum_{m=0}^{N-1} \exp(c_i \cdot c_m / \tau)} \cdot \log\left(\frac{\exp(e_i \cdot c_j / \tau)}{\sum_{m=0}^{N-1} \exp(e_i \cdot c_m / \tau)}\right),
\end{aligned}
\end{equation}
where $e$, $c$ represent the brain embedding predicted by the projector and the CLIP image embedding, respectively, and $\tau$ is the temperature hyperparameter. To enhance the quality of the reconstructed images, the reconstruction loss $\mathcal{L}_{rec.}$ is defined as follows:

\begin{equation}
\begin{aligned}
    \mathcal{L}_{rec.}(\hat{e},c)= \frac{1}{N} \sum_{i=1}^{N} \left( \hat{e}_i - c_i \right)^2,
\end{aligned}
\end{equation}
where $\hat{e}_i$ is predicted by the reconstructor.

In the new-subject few-shot adaptation stage, only the proposed FKD module is trained end-to-end using $\mathcal{L}_{total}$ to achieve brain activity visual reconstruction.
\begin{equation}
        \mathcal{L}_{total} = \mathcal{L}_{semantic} + \gamma \mathcal{L}_{Fourier},
\end{equation}
where the hyperparameter $\gamma$ is used to balance the two losses.

\section{Experiments}



\begin{table*}[h]
\centering
\caption{Quantitative comparison with SOTA methods on the NSD benchmark across all scan time settings for subject 1. Note that “–” indicates metrics not reported in the original paper.}
\footnotesize
\renewcommand{\arraystretch}{1.1} 
\begin{tabular}{c|c|cccccccc}
\toprule 
\multirow{2}{*}{Method} & \multirow{2}{*}{Scan Time} & \multicolumn{4}{c}{Low-Level}  & \multicolumn{4}{c}{High-Level}  \\ \cmidrule(lr){3-6}  \cmidrule(lr){7-10}
\makebox[0.01\textwidth][c]      &   \makebox[0.01\textwidth][c]       & \makebox[0.05\textwidth][c]{PixCorr↑}      & \makebox[0.05\textwidth][c]{SSIM↑}         & \makebox[0.01\textwidth][c]{Alex(2)↑}      &\makebox[0.01\textwidth][c]{Alex(5)↑}      &\makebox[0.01\textwidth][c]{Incep↑}        &\makebox[0.01\textwidth][c]{CLIP↑}         &\makebox[0.01\textwidth][c]{Eff↓}          &\makebox[0.01\textwidth][c]{SwAV↓}     \\ \bottomrule

Mind Reader\cite{lin2022mind}   & \multirow{5}{*}{40 h}   & -    & -    & -        & -        & 78.2\% & -      & -         & -          \\
Takagi et al. \cite{takagi2023high}  &                         & -          & -          & 83.1\%   & 83.0\%   & 75.9\% & 77.4\% &    -       & - \\
Brain-Diffuser \cite{ozcelik2023natural}  &                         & .305       & .367       & 96.7\%   & 97.4\%   & 87.8\% & 92.5\% & .768      & .415       \\
MindEye \cite{scotti2024reconstructing} &                         & .390       & .337       & 97.4\%   & 98.7\%   & 94.5\% & 94.7\% & .630      & .358       \\
\textbf{MindShot (Ours)} &                         & .321       & .285       & 93.2\%   & 96.8\%   & 94.0\% & 94.2\% & .644     & .361       \\ \hline
Takagi et al. \cite{takagi2023high}  & \multirow{4}{*}{0.36 h} & .149       & .200       & 65.1\%   & 67.6\%   & 58.0\% & 58.3\% & .966      & .695       \\
Brain-Diffuser \cite{ozcelik2023natural}  &                         & .027       & .196       & 63.8\%   & 74.0\%   & 66.7\% & 70.7\% & .919      & .581       \\
MindEye \cite{scotti2024reconstructing}  &                         & .047       & .225       & 63.8\%   & 69.8\%   & 61.4\% & 66.8\% & .925      & .596       \\
\textbf{MindShot (Ours)} &                         & .104       & \textbf{.254}       & \textbf{75.0\%}   & \textbf{83.3\%}   & \textbf{74.0\%} & \textbf{78.2\%} & \textbf{.854}      & \textbf{.533}       \\ \hline
Takagi et al. \cite{takagi2023high}  & \multirow{4}{*}{0.72 h} & .200       & .218       & 70.4\%   & 73.0\%   & 62.8\% & 61.3\% & .951      & .680       \\ 
Brain-Diffuser \cite{ozcelik2023natural}  &                         & .031          & .212          & 66.9\%          & 77.2\%          & 70.8\%          & 74.3\%          & .904          & .569          \\
MindEye \cite{scotti2024reconstructing} &                         & .079       & .233       & 69.1\%   & 76.3\%   & 68.2\% & 73.0\% & .890      & .570       \\
\textbf{MindShot (Ours)} &                         & .130 & \textbf{.257} & \textbf{78.0\%} & \textbf{87.4\%} & \textbf{80.0\%} & \textbf{83.6\%} & \textbf{.810} & \textbf{.497}       \\ \hline
Takagi et al. \cite{takagi2023high}   & \multirow{4}{*}{1.08 h} & .223       & .217       & 71.7\%   & 73.7\%   & 63.1\% & 62.2\% & .948      & .677       \\ 
Brain-Diffuser \cite{ozcelik2023natural}  &                         & .033       & .222       & 68.4\%   & 78.9\%   & 73.2\% & 76.9\% & .889      & .557       \\
MindEye \cite{scotti2024reconstructing} &                         & .083       & .252       & 72.2\%   & 79.4\%   & 70.1\% & 74.5\% & .877      & .552       \\
\textbf{MindShot (Ours)} &                         & .143 & \textbf{.261} & \textbf{80.7\%} & \textbf{89.6\%} & \textbf{82.9\%} & \textbf{84.9\%} & \textbf{.780} & \textbf{.474}       \\ \bottomrule

\end{tabular}
\label{table: few-shot}
\end{table*}

\subsection{Evaluation metrics}

To evaluate the effectiveness of our approach, several metrics are employed to assess the quality of image reconstruction. These include low-level metrics such as PixCorr, SSIM \cite{wang2004image}, AlexNet(2), and AlexNet(5) \cite{krizhevsky2012imagenet}, as well as high-level metrics such as Inception \cite{szegedy2016rethinking}, CLIP \cite{radford2021learning}, EffNet-B \cite{tan2019efficientnet}, and SwAV \cite{caron2020unsupervised}.

\subsection{Implementation details} \label{details}

All experiments are implemented in PyTorch on a single NVIDIA A100 GPU with 80GB memory. A cyclical learning rate strategy \cite{smith2017cyclical} is used, with the maximum learning rate set to $3 \times 10^{-4}$. The AdamW \cite{loshchilov2017decoupled} optimizer is employed with a weight decay of $10^{-2}$. Pretraining is performed for 240 epochs on data from three subjects, using a batch size of 64. For new subject adaptation, the batch size is 32. The Normalizer is implemented by using AdaptiveMaxPooling for dimensional normalization to ensure a uniform output length of 9600, as detailed in Section  \ref{Section: The selection of optimal hyperparameter}. The brain decoder $\mathcal{E}_{brain}$ follows a MLP architecture, consisting of a backbone, a projector, and a reconstructor. The backbone begins with a linear layer whose output size is 4096, followed by a 4-layer residual MLP, and ends with a linear layer that maps the features into $257 \times 768$. The projector is a 3-layer MLP with a hidden size of 2048. The reconstructor consists of 6 Transformer blocks with 12-head causal attention. The initial tokens are formed by concatenating the brain and image embeddings, with learnable embeddings added to the image embeddings. The prediction of image embeddings by the reconstructor is regarded as a diffusion process \cite{ramesh2022hierarchical}, progressively denoising noisy image embeddings over 100 steps. During inference, the ground truth image embeddings are masked.

Data augmentation techniques, including random cropping, horizontal flipping, color jittering, and random grayscale, are applied. During the training, one single-trial fMRI signal is randomly selected for each visual stimulus, while in the testing, the average of three fMRI signals is used. In the new-subject adaptation phase, the $\lambda$ in $\mathcal{L}_{semantic}$ is set to 30 following \cite{scotti2024reconstructing}, while $\gamma$ in $\mathcal{L}_{total}$ is set to 0.5 to balance the contributions of these losses. The $\alpha$ in $\mathcal{L}_{Fourier}$ is set to 1, and the depth of the Eliminator in the FKD module is configured as 1. The optimal hyperparameter selection is detailed in Section \ref{Section: The selection of optimal hyperparameter}. For norm setting, ridge regression \cite{hoerl1970ridge} is applied to reconstructed low-level details with 10,000 iterations and a regularization parameter of 50,000 to map the fMRI data to latent feature space, which is constructed by VAE with an image size of $512 \times 512$. Subsequently, the versatile diffusion model \cite{xu2023versatile} is employed for image reconstruction, utilizing the UniPCMultistep scheduler \cite{zhao2024unipc} with 20 steps and a guidance scale of 3.5.

\begin{table*}[h]
\centering
\caption{Quantitative comparison with the cross-subject pipeline on the NSD benchmark under extremely limited scan time.}
\footnotesize
\renewcommand{\arraystretch}{1.1} 
\begin{tabular}{c|c|c|cccccccc}
\toprule 
\multirow{2}{*}{Method} & \multirow{2}{*}{Subj} & \multirow{2}{*}{Data} & \multicolumn{4}{c}{Low-Level}  & \multicolumn{4}{c}{High-Level}  \\ \cmidrule(lr){4-7}  \cmidrule(lr){8-11}
\makebox[0.01\textwidth][c]      &   \makebox[0.01\textwidth][c]    &   \makebox[0.01\textwidth][c]   & \makebox[0.05\textwidth][c]{PixCorr↑}      & \makebox[0.05\textwidth][c]{SSIM↑}         & \makebox[0.01\textwidth][c]{Alex(2)↑}      &\makebox[0.01\textwidth][c]{Alex(5)↑}      &\makebox[0.01\textwidth][c]{Incep↑}        &\makebox[0.01\textwidth][c]{CLIP↑}         &\makebox[0.01\textwidth][c]{Eff↓}          &\makebox[0.01\textwidth][c]{SwAV↓}     \\ \bottomrule

MindEye2 \cite{scotti2024mindeye2}                        & \multirow{3}{*}{1}             & 1 h                                                                            & .235              & .428           & 88.0\%            & 93.3\%            & 83.6\%          & 80.8\%          & .798          & .459           \\
MindShot(Ours)                        &                                & 0.72 h                                                                         & .130              & .257           & 78.0\%            & 87.4\%            & 80.0\%          & 83.6\%          & .810          & .497           \\
MindShot(Ours)                         &                                & 1 h                                                                            & .142              & .267           & 82.1\%            & 90.5\%            & \textbf{83.6\%} & \textbf{85.4\%} & \textbf{.775} & .468           \\ \hline
MindEye2 \cite{scotti2024mindeye2}                        & \multirow{3}{*}{2}             & 1 h                                                                            & .200              & .433           & 85.0\%            & 92.1\%            & 81.9\%          & 79.4\%          & .807          & .467           \\
MindShot(Ours)                         &                                & 0.72 h                                                                         & .106              & .277           & 77.7\%            & 86.3\%            & 76.8\%          & 80.0\%          & .833          & .507           \\
MindShot(Ours)                         &                                & 1 h                                                                            & .127              & .287           & 80.8\%            & 89.5\%            & \textbf{83.6\%} & \textbf{84.1\%} & \textbf{.774} & \textbf{.462}  \\ \hline
MindEye2 \cite{scotti2024mindeye2}                        & \multirow{3}{*}{5}             & 1 h                                                                            & .175              & .405           & 83.1\%            & 91.0\%            & 84.3\%          & 82.5\%          & .781          & .444           \\
MindShot(Ours)                         &                                & 0.72 h                                                                         & .111              & .242           & 78.5\%            & 87.9\%            & 82.8\%          & 84.9\%          & .786          & .473           \\
MindShot(Ours)                         &                                & 1 h                                                                            & .115              & .257           & 80.8\%            & 90.2\%            & \textbf{85.8\%} & \textbf{87.1\%} & \textbf{.757} & \textbf{.447}  \\ \hline
MindEye2 \cite{scotti2024mindeye2}                        & \multirow{3}{*}{7}             & 1 h                                                                            & .170              & .408           & 80.7\%            & 85.9\%            & 74.9\%          & 74.3\%          & .854          & .504           \\
MindShot(Ours)                         &                                & 0.72 h                                                                         & .117              & .267           & 76.8\%            & 84.6\%            & 78.0\%          & 80.8\%          & .827          & .501           \\
MindShot(Ours)                         &                                & 1 h                                                                            & .127              & .272           & 78.0\%            & 86.2\%            & \textbf{79.5\%} & \textbf{82.0\%} & \textbf{.806} & \textbf{.487}  \\ \bottomrule

\end{tabular}
\label{TABLE: MindEye2}
\end{table*}

\subsection{Few-shot brain decoding performance}

\subsubsection{Quantitative comparison with SOTA methods}

As shown in Table \ref{table: few-shot}, to validate the effectiveness of MindShot, extensive quantitative comparisons are conducted with state-of-the-art following the PSPM paradigm methods \cite{takagi2023high, ozcelik2303brain, scotti2024reconstructing} on the widely used NSD dataset. Results are reported under 1-shot, 2-shot, 3-shot, and normal settings, corresponding to scan times of 0.36 h, 0.72 h, 1.08 h, and 40 h, respectively. 
In the normal setting, both high-level and low-level modules are used, whereas in the few-shot settings, only the high-level module is applied to ensure a fair comparison. The proposed MindShot method demonstrates competitive performance with SOTA methods under the normal setting, achieving a CLIP accuracy of 94.2\%. In the 1-shot scenario (scan time: 0.36 h), MindShot achieves a CLIP accuracy of 78.2\%, significantly outperforming the work \cite{takagi2023high} by 34.1\%, MindEye by 17.1\%, and Brain-Diffuser by 10.6\%. These results are consistently observed in the 2-shot and 3-shot settings. Notably, MindShot achieves a 5.0\% higher CLIP accuracy with a 0.36 h scan time, compared to MindEye's 1.08h. Furthermore, in the 2-shot scenario (scan time: 0.72 h), MindShot outperforms Takagi’s approach, which uses the entire dataset (scan time: 40 h). Extensive quantitative analysis shows that PSPM paradigm methods lack generalization and perform poorly on new subjects with limited data. In contrast, MindShot leverages cross-subject prior knowledge and the FKD model to reduce distribution differences, enabling effective adaptation to new subjects. These results highlight the potential of MindShot to advance brain decoding from traditional data-hungry models to more efficient few-shot decoding.

\begin{figure}[t]
    \centering
    \includegraphics[width=1\linewidth]{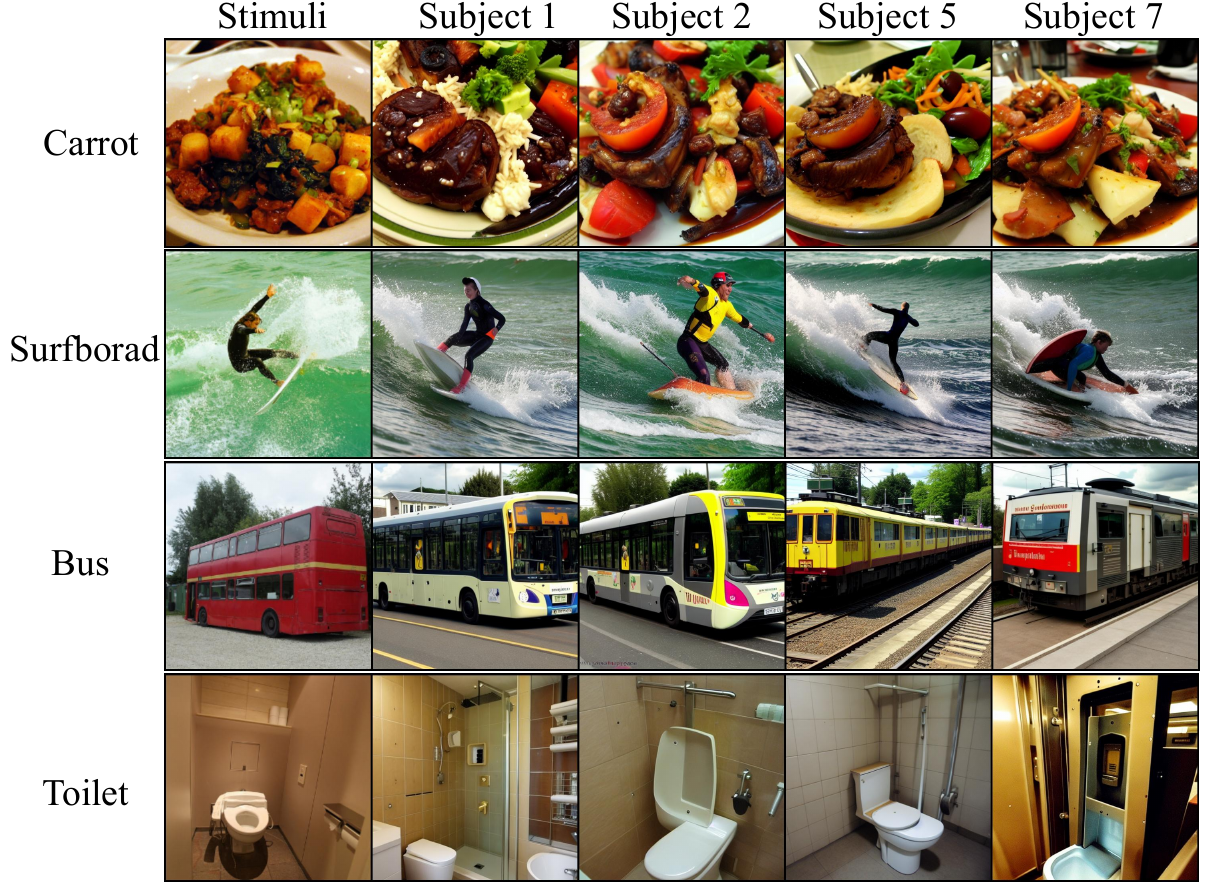}
    \caption{Visualization of one-shot brain decoding across multiple subjects, with correct recovery of food, sports, vehicles, and indoor scenes achieved under 0.36h scan time.}
    \label{fig:one-shot-visual}
\end{figure}

\subsubsection{Qualitative comparison with SOTA methods}

As shown in Fig.\ref{fig:visual}, qualitative comparisons are conducted with state-of-the-art frameworks. The first column presents the visual stimuli, followed by the reconstructed images from Takagi \textit{et al.} \cite{takagi2023high}, Brain Diffuser \cite{ozcelik2303brain}, MindEye \cite{scotti2024reconstructing}, and our method under 1-shot, 2-shot, and 3-shot settings. The visualization results indicate that Takagi \textit{et al.} fail to reconstruct meaningful images, lacking significant semantic content. Both Brain Diffuser and MindEye generate photo-like images, but their semantic information is inconsistent or ambiguous. These methods heavily rely on sufficient data approaches and perform poorly when data is scarce. These methods heavily rely on sufficient data approaches and perform poorly when data is scarce. In contrast, our proposed method leverages cross-subject supervision to achieve high-quality reconstruction with strong semantic fidelity, even under minimal data conditions.

\subsubsection{Comparison with cross-subject pipeline}

\begin{table*}[]
\centering

\caption{Ablation study of the proposed FKD module under the few-shot setting. Bold represents the best results. }
\renewcommand{\arraystretch}{1.1} 
\setlength{\tabcolsep}{3pt}
\resizebox{\linewidth}{!}{
\begin{tabular}{c|c|ccc|cccc|cccc}
\toprule
\multirow{2}{*}{Method} & \multirow{2}{*}{Few-shot} & \multirow{2}{*}{Prior} &\multirow{2}{*}{FKD} &\multirow{2}{*}{FKD-SP} & \multicolumn{4}{c|}{Low-level}                                     & \multicolumn{4}{c}{High-level}                                    \\ 
                        &                           &                        &                             &                                                                        & PixCorr↑      & SSIM↑         & Alex(2)↑        & Alex(5)↑        & Incep↑          & CLIP↑           & Eff↓          & SwAV↓         \\ \hline
Vanilla  & 1-shot   & \checkmark     &            &                                                       & .047          & .225          & 63.8\%          & 69.8\%          & 61.4\%          & 66.8\%          & .925          & .596          \\
MindShot (Scratch) & 1-shot   & \checkmark     & \checkmark          &                                                       & .086         & .251          & 72.9\%          & 82.1\%          & 72.9\%          & 77.1\%          & .866          & .538          \\
MindShot (High-level) & 1-shot   & \checkmark     & \checkmark          & \checkmark                                                     & \textbf{.104} & \textbf{.254}          & \textbf{75.0\%} & \textbf{83.3\%} & \textbf{74.0\%} & \textbf{78.2\%} & \textbf{.854} & \textbf{.533} \\ \hline
Vanilla  & 2-shot   & \checkmark     &            &                                                       & .083          & .252          & 72.2\%          & 79.4\%          & 70.1\%          & 74.5\%          & .877          & .552          \\
MindShot (Scratch) & 2-shot   & \checkmark     & \checkmark          &                                                       & .126          & .256          & 77.2\%          & 85.6\%          & 77.7\%          & 81.5\%          & .830          & .509          \\
MindShot (High-level) & 2-shot   & \checkmark     & \checkmark          & \checkmark                                                     & \textbf{.130} & \textbf{.257} & \textbf{78.0\%} & \textbf{87.4\%} & \textbf{80.0\%} & \textbf{83.6\%} & \textbf{.810} & \textbf{.497} \\ \bottomrule
\end{tabular}
}
\label{table:ablation of supervision}
\end{table*}


We compare our approach with the state-of-the-art cross-subject paradigm pipeline, namely MindEye2 \cite{scotti2024mindeye2}, which uses a linear layer to map signals from different subjects into a shared latent space, pretrains on seven subjects, and fine-tunes the entire model for new subjects. However, it does not explicitly  address inter-subject differences or explore efficient strategies for leveraging pretrained priors. In contrast, our framework pretrains on only three subjects and leverages the proposed FKD module during new-subject adaptation to account for inter-subject differences, enabling a more efficient transfer of prior knowledge. We fine-tune only the FKD module for new subjects, significantly reducing training resources while improving efficiency. As shown in Table \ref{TABLE: MindEye2}, our method significantly outperforms MindEye2 on high-level metrics across all subjects and achieves a 28\% reduction in scan time (0.72 h) while delivering competitive performance across all subjects. Notably, the proposed method outperforms MindEye2 on subject 7 by 8.7\% in the CLIP metric. These results underscore the importance of an efficient module for addressing inter-subject differences and facilitating prior knowledge transfer to mitigate data limitations in new subjects. Furthermore, MindEye2 fine-tunes SDXL \cite{podell2023sdxl} to align with the reference OpenCLIP ViT-bigG/14 image embeddings, with an additional low-level MLP submodule and reconstruction stage to refine pixel-level quality, thereby significantly improving low-level decoding performance. Given the behavioral relevance and cross-subject consistency of high-level (semantic) cortical representations, decoding high-level information is particularly important \cite{huth2016natural}. Accordingly, in cross-subject, few-shot scenarios, MindShot gives priority to high-level semantic decoding to maximize data efficiency and transferability.




\begin{table}
\caption{Performance of brain decoding across varying Eliminator depths, along with the corresponding computational overhead.}
\centering
\renewcommand{\arraystretch}{1.3} 
\resizebox{\linewidth}{!}{
\begin{tabular}{c|c|c|c|c|cccc}
\toprule 
\textbf{Depth}   &\textbf{Data} &\textbf{Para.} & \textbf{GFLOPs} & \textbf{Memory} & \textbf{Incep↑}  & \textbf{CLIP↑}         & \textbf{Eff↓}           & \textbf{SwAV↓} \\ \midrule
1     & \multirow{3}{*}{1-shot} & \textbf{92M}  & \textbf{29.59} & \textbf{10.71GB} & \textbf{74.0\%} & \textbf{78.2\%} & \textbf{.854} & \textbf{.533}  \\
2     &                         & 184M & 29.68& 11.39GB & 73.6\% & 77.1\% & .852 & .533  \\
3     &                         & 276M & 29.77& 12.08GB  & 73.8\% & 77.0\% & .852 & .536  \\ \midrule
1     & \multirow{3}{*}{2-shot} & \textbf{92M}  & \textbf{29.59} & \textbf{10.71GB}  & \textbf{80.0\%} & \textbf{83.6\%} & \textbf{.810} & \textbf{.497}  \\
2     &                         & 184M & 29.68& 11.39GB  & 79.8\% & 82.9\% & .812 & .498  \\
3     &                         & 276M & 29.77& 12.08GB  & 79.3\% & 82.7\% & .810 & .498  \\ 

\bottomrule
\end{tabular}
}
\label{table: Depth}
\end{table}

\begin{figure}[]
    \centering
    \includegraphics[width=1\linewidth]{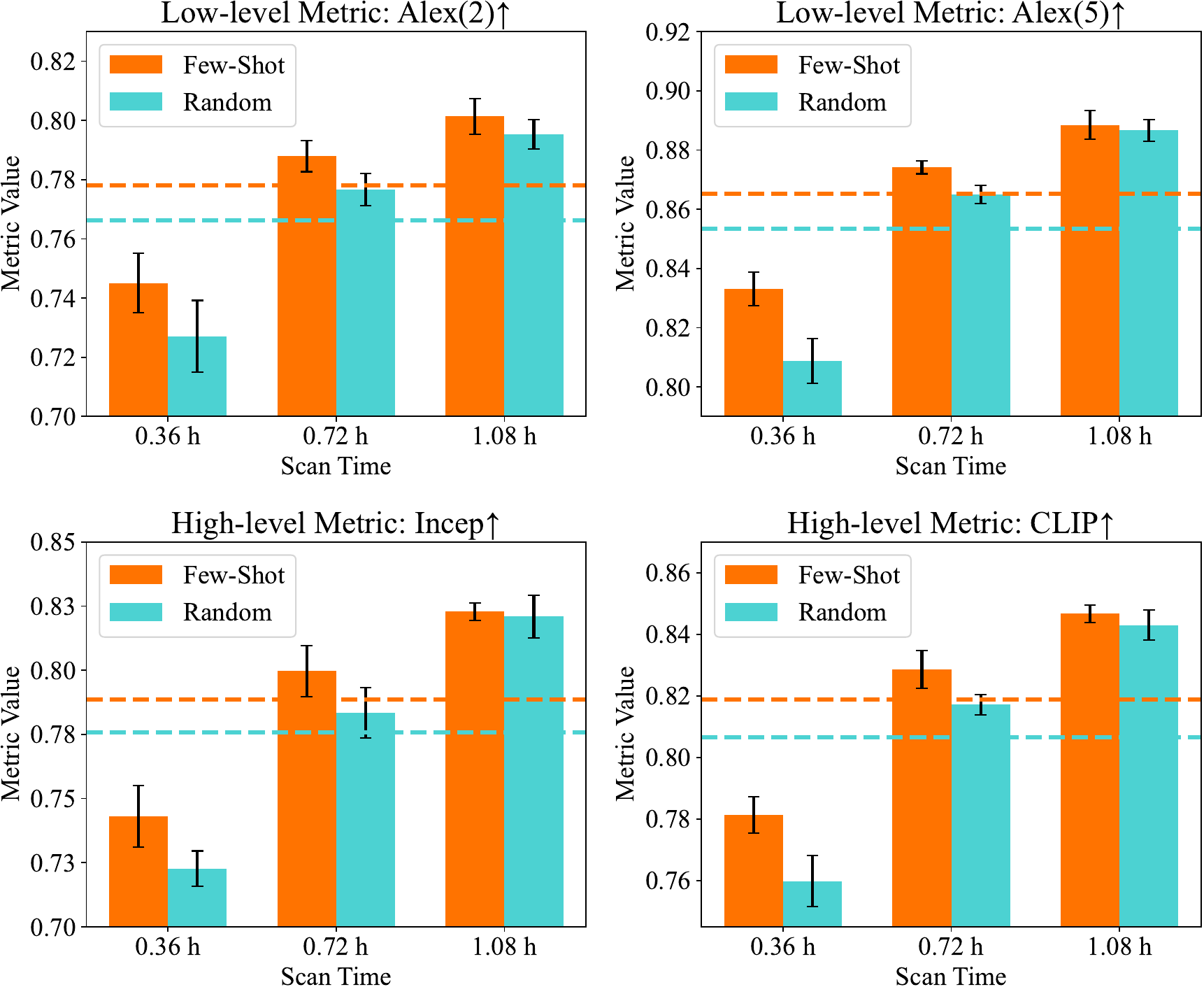}
    \caption{Brain decoding performance comparison of MindShot under two sampling methods: few-shot and random sampling. The dashed line represents the average score across all scan times.}
    \label{fig:shot setting}
\end{figure}

\subsubsection{Qualitative results on one-shot brain decoding}

\begin{figure*}[t]
    \centering
    \includegraphics[width=0.95\linewidth]{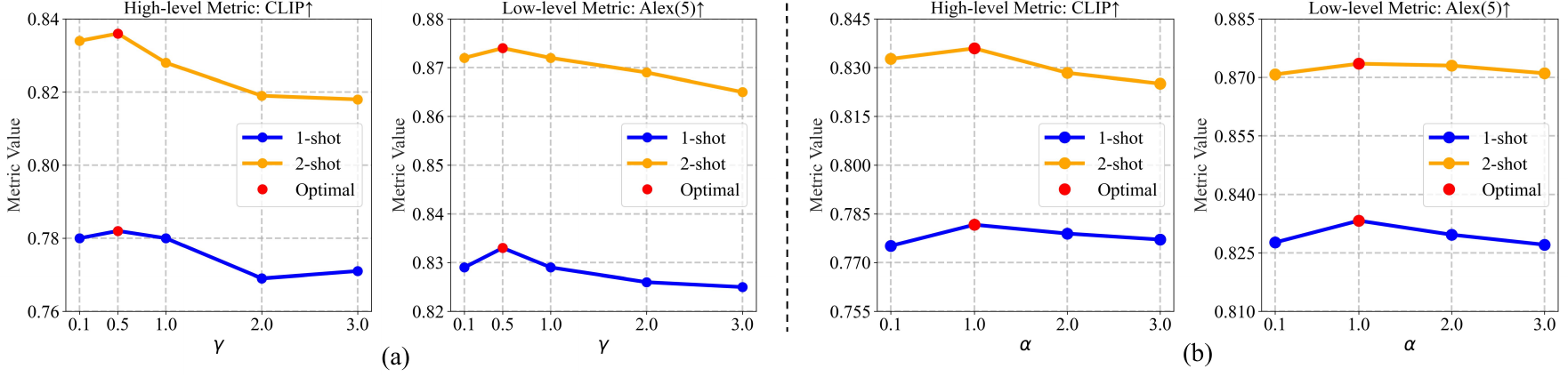}
    \caption{
    Ablations. (a) Decoding performance under varying values of $\gamma$ in $\mathcal{L}_{total}$ across different scenarios. (b) Decoding performance under varying values of $\alpha$ in $\mathcal{L}_{Fourier}$ across different scenarios.}
    \label{fig: ablation floss and weighting alpha}
\end{figure*}

As shown in Fig.\ref{fig:one-shot-visual}, our method performs brain decoding in the one-shot setting (scan time: 0.36h), demonstrating strong generalization by successfully decoding across multiple new subjects. The reconstructed images also demonstrate high diversity, covering concepts such as food, sports, vehicles, indoor scenes, etc. Specifically, in the first row for food-related stimuli, MindShot captures intricate details and produces high-quality, richly colored reconstructions. In the second row, representing outdoor sports scenes, MindShot accurately reconstructs key semantic features, such as a surfboard, and captures dynamic human actions and waves. In the third row, for vehicle-related stimuli, subjects 1 and 2 successfully generate images of a bus category, while subjects 5 and 7 reconstruct images resembling a train. This difference arises due to the visual similarity between buses and trains. In the final row, depicting indoor scenes, the key elements of a toilet scene are reconstructed for all subjects. Overall, despite the scan time being only 0.36 h in the one-shot setting, MindShot demonstrates strong generalization across multiple subjects, high-quality reconstructions, and semantic diversity, indicating promising potential for practical applications.

In summary, the proposed MindShot addresses three key challenges: (1) significantly decreasing cross-subject individual differences, (2) efficiently utilizing cross-subject prior knowledge to tackle the issue of limited samples in the few-shot setting, and (3) achieving the visual reconstruction through a cross-modal approach that ensures high semantic fidelity and fine-grained details.

\subsection{Ablation study}

\begin{figure}[t]
    \centering
    \includegraphics[width=0.75\linewidth]{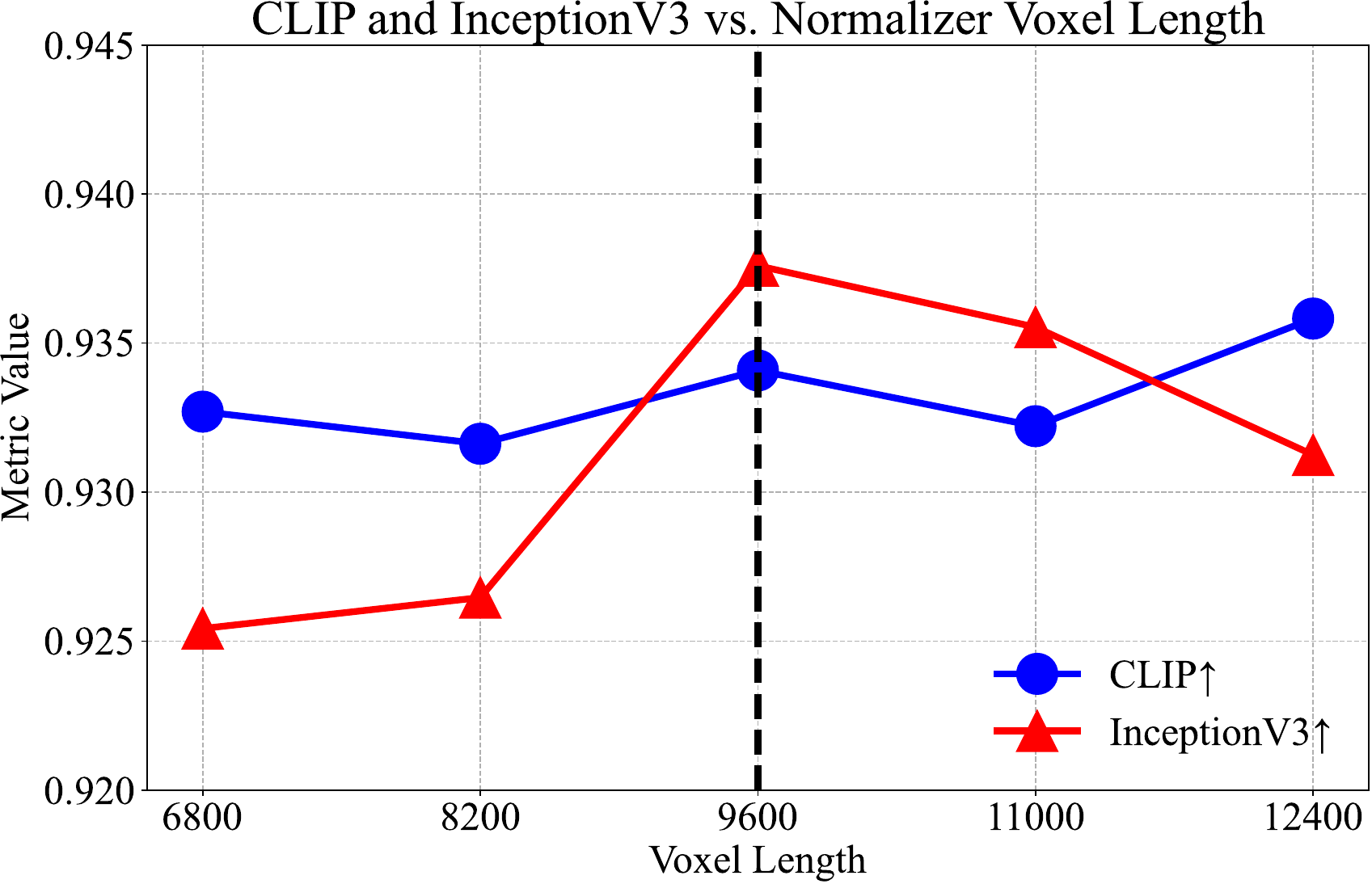}
    \caption{
    Brain decoding performance with different Normalizer voxel length outputs.}
    \label{fig: ablation_voxel_length.pdf}
\end{figure}


\subsubsection{Effectiveness of the proposed FKD module}

To validate the effectiveness of the proposed method, as shown in Table \ref{table:ablation of supervision}, extensive experiments are conducted under few-shot settings on subject 1. In the first row of the table, the Vanilla method, trained with only 1-shot (0.36 h) and pretrained with cross-subject prior knowledge, results in a CLIP accuracy of 66.8\%. Due to the limited data for new subjects, large individual differences interfere with the pretrained model. In contrast, the proposed FKD module aligns signal distribution across individuals in the frequency domain, which enables more efficient use of cross-subject priors and improves the CLIP score by 15.4\%. The same improvement is observed in the 2-shot setting. Furthermore, as shown by MindShot (High-level), full-scale training during new subject adaptation may lead to catastrophic forgetting and performance deterioration. Therefore, we train only a subject-specific, plug-and-play FKD (FKD-SP) module during new subject adaptation while keeping the backbone frozen. This approach not only improves the CLIP metric to 78.2\% but also preserves the backbone's performance across multiple subjects.

\subsubsection{Effectiveness of the few-shot setting}

As shown in Fig.\ref{fig:shot setting}, to validate the effectiveness of the few-shot setting under conditions of data scarcity, we conduct a comprehensive comparison between the few-shot setting and random sampling at scan times of 0.36 h, 0.72 h, and 1.08 h, each repeated five times with different random seeds. The results show that the proposed few-shot setting consistently outperforms the random sampling method across all scan times in terms of average performance. Notably, under the most limited data condition (0.36 h), the few-shot setting achieves the largest improvement, highlighting the importance of effective sampling strategies when data are scarce.

\subsubsection{The selection of optimal hyperparameter} \label{Section: The selection of optimal hyperparameter}



\begin{figure*}[t]
    \centering
    \includegraphics[width=1\linewidth]{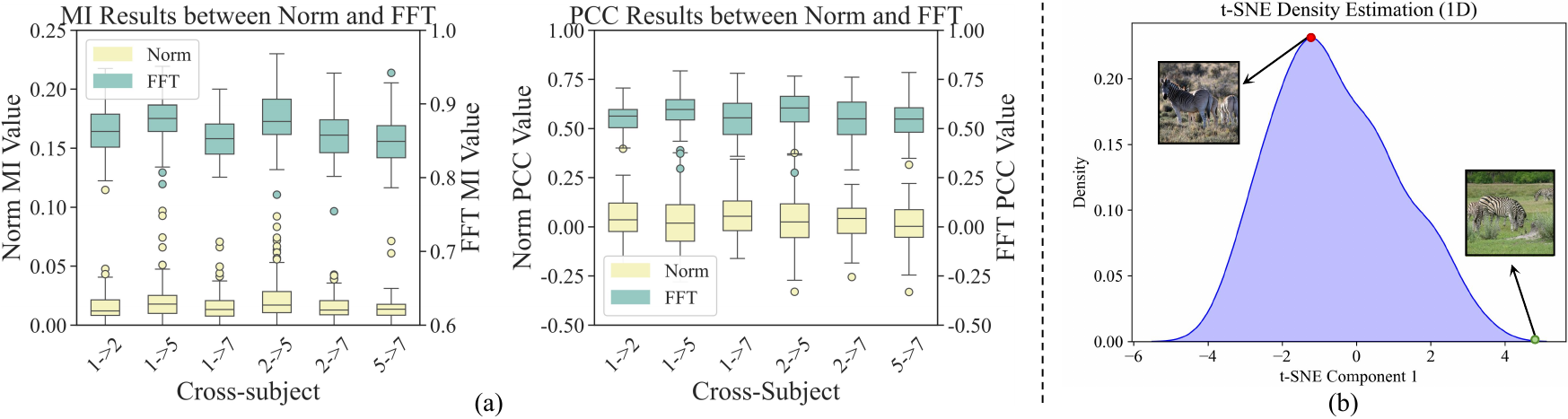}
    \caption{Discussions. (a) Statistical results of cross-subject similarity comparison under the Norm and FFT methods. (b) Visualization of one-shot selection. Taking the zebra semantic class as an example, the red and green dots represent the visual stimulus with the highest and lowest probability density, respectively.}
    \label{fig: FFT_norm_and tsne.pdf}
\end{figure*}

To provide more practical guidance and examine the sensitivity of MindShot to key hyperparameters, including $\gamma$ in $\mathcal{L}_{total}$, $\alpha$ in $\mathcal{L}_{Fourier}$, the depth of the Eliminator, and the output size of the Normalizer, we provide their configurations below.

\textbf{$\gamma$}: The $\gamma$ in $\mathcal{L}_{total}$ is used to balance semantic accuracy for new individuals and reduce inter-individual variability. As shown in Fig.\ref{fig: ablation floss and weighting alpha} (a), decoding performance is evaluated using the CLIP and the Alex(5) metrics under 1-shot and 2-shot settings across different values of $\gamma$. The results show that $\gamma = 0.5$ is the optimal choice.

\textbf{$\alpha$}: The $\alpha$ in $\mathcal{L}_{Fourier}$ is used to balance the contributions of phase and amplitude components. As shown in Fig.\ref{fig: ablation floss and weighting alpha} (b), the optimal value of $\alpha$ is 1.0, indicating that both components play an equal role in mitigating individual variability.

\textbf{Depth of Eliminator:} The depth determines FKD’s capacity to reduce inter-subject differences when adapting to a new subject. As shown in Table \ref{table: Depth}, the performance is reported across two data scales and varying depths. The depth-1 configuration achieves optimal performance while maintaining the lowest computational overhead.

\textbf{Output Size of Normalizer:} The Normalizer serves two primary functions: reducing absolute variability and aligning signal dimensions across subjects, and compressing inputs while retaining informative components. In the MSP stage, we conduct preliminary experiments to determine an appropriate output size. As shown in Fig.\ref{fig: ablation_voxel_length.pdf}, the decoding performance based on CLIP and Incep. metrics is reported across output sizes ranging from 6,800 to 12,400, with the optimal value identified as 9,600.

\section{Discussion}


\subsection{How FKD module enhances few-shot brain decoding performance}

\begin{table}[t]
\centering
\caption{Cross-subject similarity comparison between the Normal and FFT methods across different subject pairs.}
\renewcommand{\arraystretch}{1.3} 
\resizebox{\linewidth}{!}{
\setlength{\tabcolsep}{4pt} 
\begin{tabular}{c|c|cc|c|c|cc}
\toprule
\multicolumn{2}{c|}{Method}                       & MI   & PCC  & \multicolumn{2}{c|}{Method}                       & MI   & PCC  \\ \midrule
\multirow{2}{*}{Subj 1-\textgreater{}2} & Normal & .016 & .038 & \multirow{2}{*}{Subj 2-\textgreater{}1} & Normal & .019 & .047 \\
                                        & FFT    & .865 & .557 &                                         & FFT    & .884 & .594 \\ \hline
\multirow{2}{*}{Subj 1-\textgreater{}5} & Normal & .021 & .027 & \multirow{2}{*}{Subj 5-\textgreater{}1} & Normal & .018 & .007 \\
                                        & FFT    & .881 & .589 &                                         & FFT    & .880 & .589 \\ \hline
\multirow{2}{*}{Subj 1-\textgreater{}7} & Normal & .016 & .056 & \multirow{2}{*}{Subj 7-\textgreater{}1} & Normal & .020 & .045 \\
                                        & FFT    & .854 & .552 &                                         & FFT & .878 & .591 \\ \bottomrule
\end{tabular}
}
\label{Table: MI Normal}
\end{table}

In this section, we take a further step to analyze, from both theoretical and experimental perspectives, why the FKD module is better at eliminating inter-subject differences.

Firstly, as mentioned earlier, the phase and amplitude components of the signal spectrum correspond to high-level semantics and low-level statistical details, respectively. The FKD module is designed to focus on the relevant, meaningful components of the signal in the frequency domain, while discarding or minimizing the impact of other noise.

Secondly, as shown in Table \ref{Table: MI Normal}, we use mutual information (MI) \cite{cover1999elements} and Pearson correlation coefficients (PCC) to measure the inter-subject similarity of signals corresponding to the same semantic visual stimuli. We sample 2-shot examples, where one shot is used for training and the other for testing. Ridge regression is applied to both the original signals (Norm) and the FFT for cross-subject training. In the first row, for the original signals (Norm) between subjects 1 and 2, the MI and PCC values are 0.016 and 0.038, respectively. After applying FFT, the correlation increases to 0.865 and 0.557, respectively. Similar results are observed for the pairs $2 \to 1$, $1 \to 5$, $5 \to 1$, etc. This demonstrates that the proposed FKD module, by providing supervision in the frequency domain, helps mitigate individual differences and facilitates the transfer of prior knowledge.

Moreover, as shown in Fig.\ref{fig: FFT_norm_and tsne.pdf} (a), we present the MI and PCC metrics before (Norm) and after FFT in a cross-subject setting with subjects 1, 2, 5, and 7. The left plot shows the MI metric, while the right plot shows the PCC metric. We observe that (1) in the frequency domain, individual differences are reduced, thus facilitating better utilization of cross-subject prior knowledge, (2) the cross-subject combinations $1 \to 5$ and $2 \to 5$ outperform others in both MI and PCC metrics, indicating that subject 5 has high transferability and benefits more from knowledge transfer. This is further supported by Table \ref{TABLE: MindEye2}, where both MindEye2 \cite{scotti2024mindeye2} and MindShot (Ours) achieve optimal decoding performance for subject 5. Specifically, with just 0.72 hours of scan time, MindShot achieves a CLIP score of 84.9\%, surpassing subjects 1, 2, and 7. These results show that despite individual differences, certain individuals exhibit stronger generalizability, which enhances the overall performance of cross-subject decoding.


\subsection{Few-shot visual stimuli selection} \label{Exp:Selection}

\begin{figure}[t]
    \centering
    \includegraphics[width=1\linewidth]{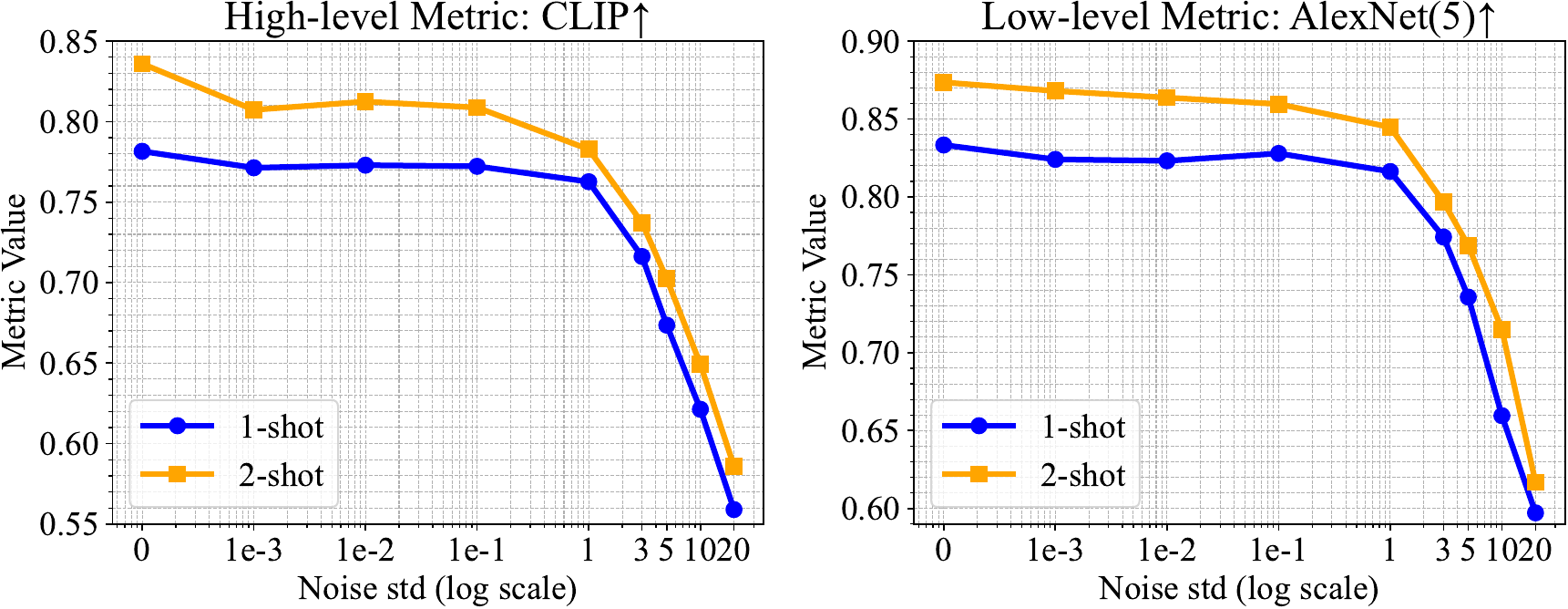}
    \caption{Brain decoding performance under varying levels of Gaussian noise (different standard deviations).}
    \label{fig: noised_mindshot.pdf}
\end{figure}


MindShot serves as a validation framework for selecting the most suitable samples in the few-shot setting by analyzing the statistical distribution of each semantic category. Specifically, for the zebra category, as shown in Fig.\ref{fig: FFT_norm_and tsne.pdf}, the fMRI voxels for all zebra-related images are reduced to one dimension based on the t-SNE method \cite{van2008visualizing}. A Gaussian distribution is used to estimate the probability density function, thus enabling the selection of images with the highest and lowest density. As shown in Table \ref{table:Selection}, the max-density method represents the core of an individual’s distribution and exhibits significant differences from other individuals, which hinders cross-subject knowledge transfer and achieves the lowest CLIP score of 75.0\%. In contrast, the min-density method samples fMRI signals farther from the central distribution, exhibiting higher transferability and achieving a CLIP score of 77.7\%. However, the inherent noise introduced by imaging equipment may affect model training, thus leading to suboptimal decoding performance. Consequently, random selection within semantic clusters achieves a balance between individual variability and noise, delivers the highest decoding performance with a CLIP score of 78.2\%.


\begin{figure}[t]
    \centering
    \includegraphics[width=1\linewidth]{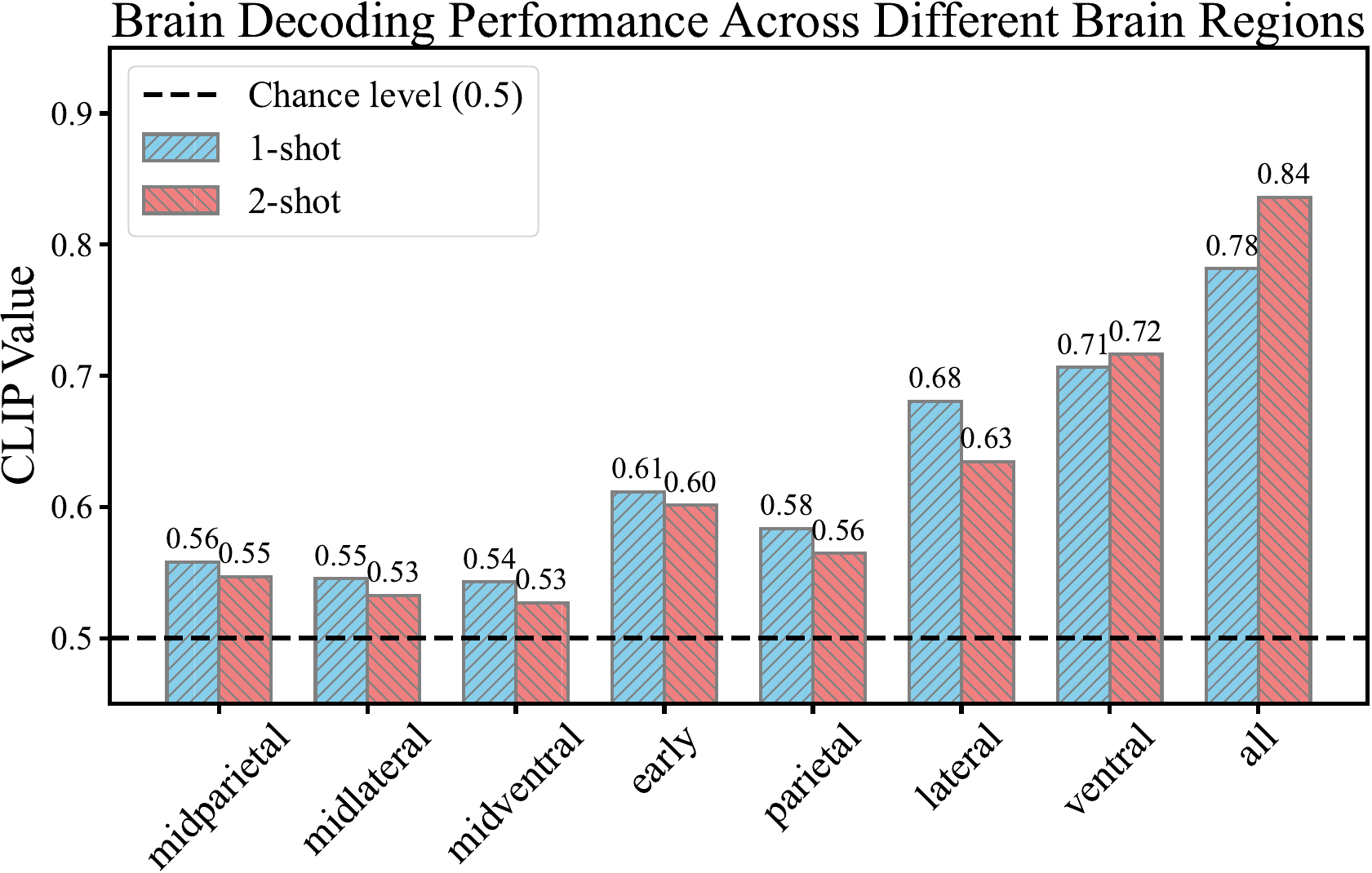}
    \caption{
    Brain decoding performance across different ROI masks.
    }
    \label{fig: roi_mask.pdf}
\end{figure}

\begin{table}
\caption{Brain decoding results for one-shot on three visual stimulus sampling strategies.}
\centering
\renewcommand{\arraystretch}{1.1} 
\resizebox{\linewidth}{!}{
\begin{tabular}{cccccc}
\toprule 
\textbf{Method}   &\textbf{Data} & \textbf{Incep↑}  & \textbf{CLIP↑  }         & \textbf{Eff↓}           & \textbf{SwAV↓} \\ \midrule
Max-density      &1-shot  & 70.6\% & 75.0\%         & 0.867          & 0.547 \\
Min-density    &1-shot   & 74.0\%  & 77.7\% & 0.859          & 0.544 \\ 
Random         &1-shot   & \textbf{74.0\%}   & \textbf{78.2\%}         & \textbf{0.854}          & \textbf{0.533} \\ \hline
Max-density      &2-shot  & 78.0\% & 81.2\%         & 0.819          & 0.505 \\
Min-density    &2-shot   & 79.6\%  & 83.4\% & 0.815          & 0.501 \\ 
Random         &2-shot   & \textbf{80.0\%}   & \textbf{83.6\%}         & \textbf{0.810}          & \textbf{0.497} \\

\bottomrule
\end{tabular}
}
\label{table:Selection}
\end{table}

\subsection{Clinical Applicability} \label{Exp:Clinical Applicability}

Beyond its competitive performance in shortened scanning times, MindShot demonstrates the potential to be integrated into clinical and personalized neuroimaging pipelines from three perspectives.

Firstly, the proposed plug-and-play FKD module enables rapid adaptation to individual brain patterns with minimal scanning, while preserving the prior knowledge learned from multi-subject data. This allows MindShot to achieve both subject-specific customization and cross-subject generalization.

Secondly, considering the inherent challenges in clinical environments, such as medical constraints and comorbidities, the quality of brain signals often suffers from various uncertainties. As shown in Fig.\ref{fig: noised_mindshot.pdf}, simulation experiments are conducted by introducing Gaussian noise (mean = 0, standard deviation ranging from 1e-3 to 20) to the Z-score normalized fMRI signals. The results show that when std $\leq$ 1, CLIP accuracy drops by only 2.4\% under one-shot, and MindShot still remains competitiveness even when std = 3 or 5, demonstrating its potential for deployment in challenging clinical environments with compromised signal quality.

Furthermore, according to the cortical parcellation in \cite{wang2015probabilistic}, a detailed analysis is conducted on seven functional cortical regions, namely early, midparietal, midlateral, midventral, parietal, lateral, and ventral. As illustrated in Fig.\ref{fig: roi_mask.pdf}, the decoding performance is evaluated under various few-shot settings using different ROI masks. Experimental results reveal that the early, lateral, and ventral regions achieve accuracies of 0.61, 0.68, and 0.71 while using only 37.7\%, 15.1\%, and 23.1\% of the original scanning range. Notably, the ventral region retains 91\% of the full-voxel decoding performance, underscoring its critical role in high-level semantic decoding. This phenomenon is consistent with the established role of the ventral visual pathway in face recognition and object semantic understanding \cite{grill2014functional}, providing the evidence for the potential of MindShot to narrow scanning scope in scenarios with limited equipment or visual impairments.

\begin{figure}[t]
    \centering
    \includegraphics[width=1\linewidth]{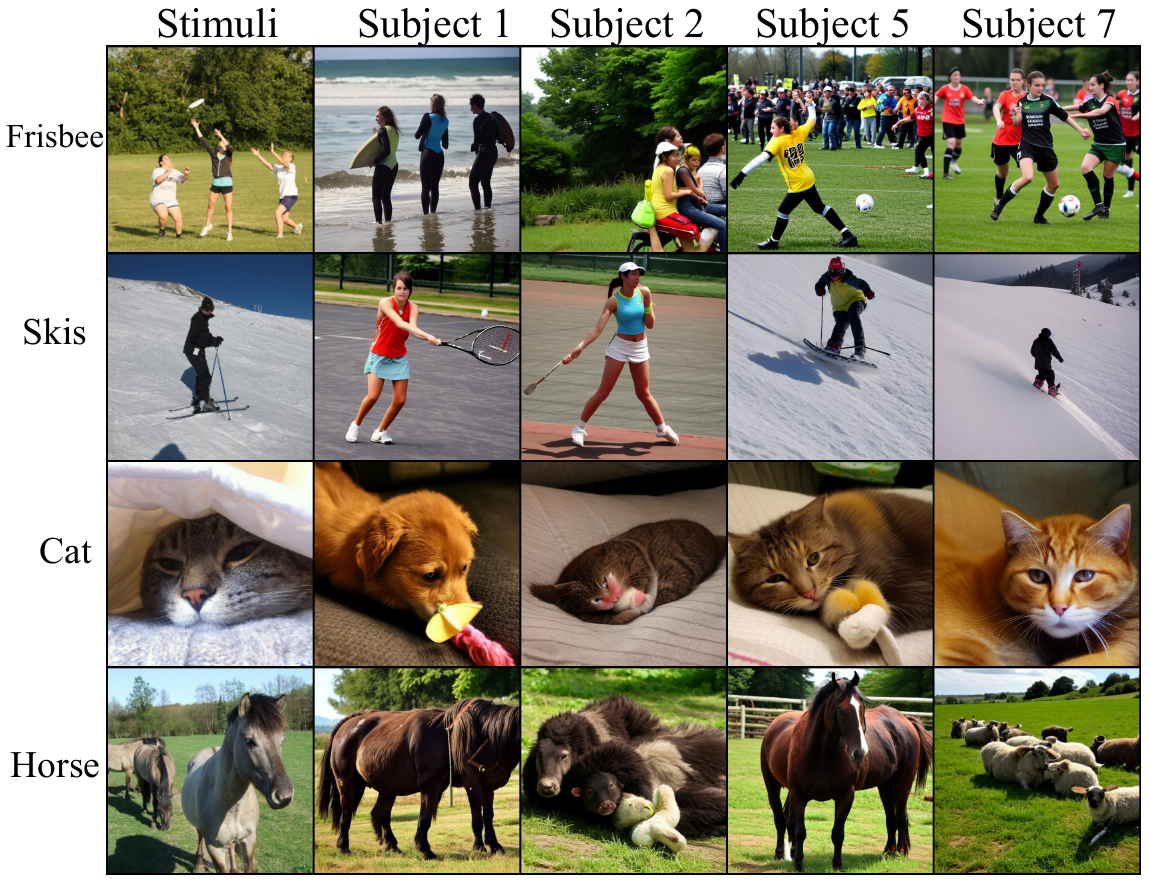}
    \caption{
    Decoding inconsistencies among subjects for hard samples from semantic categories that closely related.
    }
    \label{fig: bad_cases.pdf}
\end{figure}

\subsection{Limitations and reproducibility} 

 MindShot has limitations in distinguishing closely related semantic categories and cannot always ensure consistent decoding across subjects under such circumstances. As shown in Fig.\ref{fig: bad_cases.pdf}, subjects or model may yield incorrect results—for example, decoding the \textit{cat} as the \textit{dog} or the \textit{horse} as the \textit{cow} or \textit{sheep}. Moreover, the current fMRI–image dataset \cite{allen2022massive} lacks adequate subject diversity and paired low-field–scanner acquisitions. To advance clinical adaptation, future work could evaluate the decoding approach with more diverse participants, fewer or even zero-shot scans, lower-quality scanners, and mental-imagery tasks \cite{kneeland2025nsd}.

 Furthermore, as shown in Algorithm \ref{Algorithm:FKD}, we provide additional pseudocode for the FKD module. And the complete implementation of MindShot is publicly available at https://github.com/JSinBUPT/MindShot.

\begin{algorithm}[t]
\captionsetup{font=footnotesize} 
\small
\caption{Fourier-based Cross-subject Knowledge Distillation.}
\textbf{Input:} New subject signal $x_i$, pretrained subjects' signals $X$\\
\textbf{Output:} Distilled signal $\hat{x}_i $, $\mathcal{L}_{Fourier}$
\begin{algorithmic}[1]
\STATE Sample cross-subject signals: $x_j \gets \text{Sampler}(X)$ 
\STATE Normalize both signals: \\ $x_i \gets \text{Normalizer}(x_i)$, $x_j \gets \text{Normalizer}(x_j)$ 
\STATE Residual-based Eliminator: $\hat{x}_i \gets \text{Linear}(x_i) + x_i$
\STATE Fourier transform Eq.\ref{Eqa: F}: \\ $\mathcal{F}(\hat{x}_i) \gets \text{FFT}(\hat{x}_i)$, $\mathcal{F}(x_j)\gets \text{FFT}(x_j)$ 
\STATE Decompose amplitude and phase Eq.\ref{Eqa AP}: \\ $\mathcal{A}_i$,$\mathcal{P}_i \gets |\mathcal{F}(\hat{x}_i)|$,$\angle \mathcal{F}(\hat{x}_i)$; $\mathcal{A}_j$,$\mathcal{P}_j \gets |\mathcal{F}(x_j)|$,$ \angle \mathcal{F}(x_j)$
\STATE Compute amplitude and phase losses Eq.\ref{Eqa lamp lpha}: \\ $\mathcal{L}_{amp} \gets \text{MSE}(\mathcal{A}_i, \mathcal{A}_j)$, $\mathcal{L}_{pha} \gets \text{MSE}(\mathcal{P}_i, \mathcal{P}_j)$
\STATE Obtain the Fourier loss Eq.\ref{Eqa Fourier}: $\mathcal{L}_{Fourier} \gets \mathcal{L}_{pha} + \lambda \cdot \mathcal{L}_{amp}$
\STATE \textbf{return} $\hat{x}_i, \mathcal{L}_{Fourier}$
\end{algorithmic}
\label{Algorithm:FKD}
\end{algorithm}

\section{Conclusions}

In this paper, we introduce MindShot, a few-shot brain decoding framework designed to address the challenges of data scarcity and individual differences. First, MindShot employs multi-modal contrastive learning to align fMRI data from multiple subjects into a shared semantic space, thereby acquiring cross-subject prior knowledge. Second, the proposed FKD module, equipped with the Eliminator and $\mathcal{L}_{Fourier}$, reduces individual differences in the frequency domain. Finally, under few-shot settings, MindShot enables efficient knowledge transfer and achieves high-quality image reconstruction with the assistance of a generative model. Notably, on the NSD dataset, MindShot achieves a CLIP accuracy of 83.6\% with only 0.72 h of scan time, outperforming the PSPM methods that require 40 h of training data to reach 77.4\% accuracy. The results demonstrate not only the feasibility of few-shot brain decoding but also the potential for training large-scale models with reduced data dependence.

\section{Declaration of generative AI in scientific writing}

The authors don't use generative AI for writing and are fully responsible for the content of the paper.

\section{CRediT authorship contribution statement}

\textbf{Shuai Jiang:} Conceptualization, Data curation, Formal analysis, Investigation, Methodology, Resources, Software, Validation, Visualization, Writing - original draft. \textbf{Zhu Meng:} Funding acquisition, Methodology, Validation, Writing - review \& editing. \textbf{Haiwen Li:} Conceptualization, Methodology, Validation, Writing – review \& editing. \textbf{Delong Liu:} Validation, Visualization, Writing - review \& editing. \textbf{Fei Su:} Funding acquisition, Project administration, Resources, Supervision. \textbf{Zhicheng Zhao:} Conceptualization, Methodology, Project administration, Supervision, Validation, Visualization, Writing - review \& editing.

\section{Acknowledgments}
This work is supported by Chinese National Natural Science Foundation (62401069).

\bibliographystyle{elsarticle-harv}  
\biboptions{sort&compress}
\bibliography{kbs2025references}

\end{document}